\documentclass[letterpaper, 10 pt, conference]{ieeeconf}  

\IEEEoverridecommandlockouts                              

\usepackage{amsmath, amssymb} 
\usepackage{bbold}
\usepackage{wrapfig}
\usepackage{subfigure}
\usepackage{siunitx}
\usepackage{booktabs}
\usepackage{graphicx} 
\usepackage[font=footnotesize,labelfont=bf]{caption}
\usepackage{pgf}
\usepackage{bm}

\newcommand*\diff{\mathop{}\!\mathrm{d}}

\title{\LARGE \bf
Learning Robotic Manipulation Skills Using an Adaptive Force-Impedance Action Space
}

\author{Maximilian Ulmer, Elie Aljalbout, Sascha Schwarz, and Sami Haddadin
\thanks{All authors are with the Technical University of Munich (TUM), 80797 Munich, Germany 
        {\tt\small \{name.lastname@tum.de\}}} %
}

\begin{document}

\maketitle
\thispagestyle{empty}
\pagestyle{empty}

\begin{abstract}
Intelligent agents must be able to think \textit{fast} and \textit{slow} to perform elaborate manipulation tasks. Reinforcement Learning (RL) has led to many promising results on a range of challenging decision-making tasks. However, in real-world robotics, these methods still struggle, as they require large amounts of expensive interactions and have \textit{slow} feedback loops. On the other hand, \textit{fast} human-like adaptive control methods can optimize complex robotic interactions, yet fail to integrate multimodal feedback needed for unstructured tasks. In this work, we propose to factor the learning problem in a hierarchical learning and adaption architecture to get the best of both worlds. The framework consists of two components, a slow reinforcement learning policy optimizing the task strategy given multimodal observations, and a fast, real-time adaptive control policy continuously optimizing the motion, stability, and effort of the manipulator. We combine these components through a bio-inspired action space that we call AFORCE. We demonstrate the new action space on a contact-rich manipulation task on real hardware and evaluate its performance on three simulated manipulation tasks. Our experiments show that AFORCE drastically improves sample efficiency while reducing energy consumption and improving safety. 
\end{abstract}

\section{Introduction}
Deep reinforcement learning (RL) is a promising approach to solve complex robotic manipulation tasks by enabling seamless integration of perception, decision-making, and control. Such methods can learn mappings from a variety of sensors to actions but may require millions of interactions to converge, especially in the real world~\cite{barth2018distributed}. This sample inefficiency could be attributed to the inherent properties of the action and observation space of many real-world tasks. 

Real-world action spaces are inherently continuous. Yet, finding the correct action only solves part of the problem. To achieve the desired outcome of an action, the agent must control the interaction appropriately. In the context of manipulation, the agent must be able to compensate for any forces arising from physical interaction. Moreover, it must have the ability to continuously adapt the mechanical impedance at contact points to stabilize the manipulation~\cite{burdet2001central, won1995stability}. Such real-time adaption is important, however, most modern deep RL methods fail to achieve it due to feedback loops being slower than the robot control frequency.

Real-world observation spaces are inherently multimodal. In addition, different sensor signals must be evaluated at different time intervals. Visual information often changes slowly over time and requires deliberate reasoning to process. In contrast, proprioceptive measurements need to be evaluated in real-time for control and sometimes require an unconscious, intuitive response to establish reactive and safe behavior~\cite{haddadin2008collision}. In recent work, RL has shown to allow elegant integration of multiple modalities~\cite{lee2019making, zambelli2021learning}, yet, it remains challenging to account for temporal context.

In this work, we propose to combine bio-inspired adaptive control and reinforcement learning. Inspired by studies in neuroscience~\cite{botvinick2009hierarchically, kahneman2011thinking, ikegami2021hierarchical}, we model the solution of learning robotic manipulation as a hierarchical control policy. An outer high-level policy -- or task planner -- produces actions given observations of the environment. It is optimized in an RL framework through the maximization of future reward~\cite{sutton2018reinforcement}. Additionally, an inner low-level policy computes robot commands given high-level actions and plant measurements. It optimizes the interaction between the manipulator and its environment by minimizing instability, motion error, and effort~\cite{yang2011human}. 

The key idea of this work is to enable modular integration of perception modules and modern control paradigms, yet reducing the computational complexity of the individual components. Intuitively, the RL policy is not required to optimize the interaction with the environment through tactile and proprioceptive sensing, hence we keep the RL observation and action space small. 
\begin{figure}[t]
    \centering
    \includegraphics[width=0.4\textwidth]{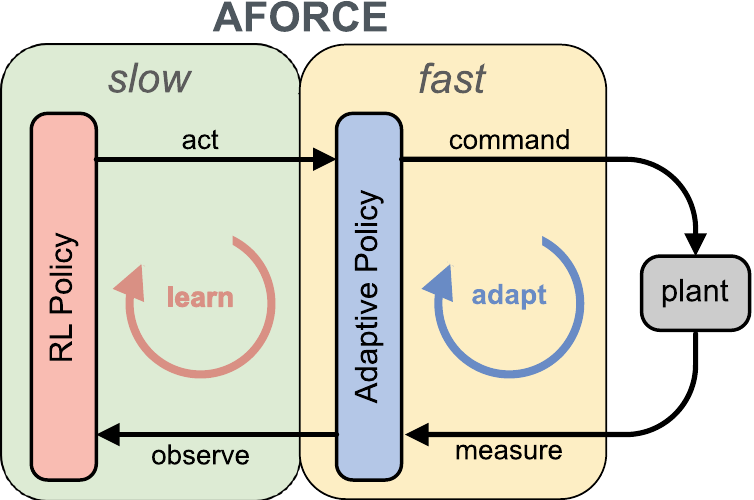}
    \caption{Intelligent agents have to think fast and slow to solve complex manipulation tasks. AFORCE combines \textit{slow} reinforcement learning with \textit{fast} adaptive control to get the best of both worlds.}
    \label{fig:graph_abstract}
    \vspace{-2mm}
\end{figure}

To combine these methods, we propose Cartesian \underline{A}daptive \underline{For}ce-Impedan\underline{ce} Control (AFORCE) as a novel action space for learning robotic manipulation. AFORCE allows continuous adaption of impedance and force, both critical to contact stability, disturbance compensation, and energy efficiency. We show that AFORCE significantly increases the sample efficiency and stability of RL agents learning manipulation tasks. In addition, we demonstrate the efficacy of the action space on real hardware by performing a contact-rich manipulation task and comparing its energy efficiency to other compliant action spaces. In this work, we focus on validating the action space and leave vision-based RL experiments for future work.

Our primary contributions are: (i) AFORCE, a bio-inspired action space implementing a hierarchical control policy for learning contact-rich manipulation tasks, (ii) A perspective on human motor control and its capacity to improve RL policies, (iii) Evaluation of the learning efficiency and comparison to baseline action spaces in robotic manipulation tasks, and (iv) Demonstration of the action space efficacy in real-world robot experiments.

\section{Related Work}
\label{sec:related_work}
\textbf{Adaptive Impedance Control} is a paradigm that takes inspiration from human motor control by adapting endpoint force and impedance to compensate for environment forces and instability~\cite{franklin2007endpoint, franklin2008cns}. The dynamic properties of such controllers were previously analyzed and demonstrated in~\cite{yang2011human}. In~\cite{ganesh2010biomimetic} the paradigm is extended to iteratively adjust a preplanned trajectory in the presence of obstacles. Furthermore, ~\cite{li2018force} introduce a trajectory adaption law, with the assumption that the desired contact force is maintained and a reference trajectory exists. \cite{johannsmeier2019framework} propose a framework to optimize adaptive parameters, given a trajectory that can solve the task. Our method builds on the control laws introduced and analyzed in~\cite{ganesh2010biomimetic, yang2011human}. However, in contrast to prior work, we don't optimize the task trajectory in the same process as force and impedance or assume the trajectory to be given. Instead, we propose to factor the system in a hierarchical control structure such that the task trajectory can be learned with reinforcement learning.

\textbf{Hierarchical Reinforcement Learning (HRL).} ~\cite{botvinick2009hierarchically, botvinick2012hierarchical} is widely used for robotic tasks, such as manipulation~\cite{beyret2019dot}, navigation~\cite{bischoff2013hierarchical}, and locomotion~\cite{jain2019hierarchical}. In this work, we refrain from learning the low-level policy parameters, hence our method is not HRL-based. Nonetheless, AFORCE relies on a hierarchical control structure at its core, hence, we briefly survey advances in the field to better position our work. Many HRL algorithms have been proposed, such as Hierarchies of Machines~\cite{parr1998reinforcement}, MAXQ~\cite{dietterich2000hierarchical}, Hierarchical Policy Search \cite{ghavamzadeh2003hierarchical}, and more recently LaDiPS~\cite{end2017layered}. Such a structure allows temporal and state abstractions which reduces the computational complexity of the individual components. In this work, we use this concept with a model-free high-level RL policy and a traditional model-based low-level policy, both containing learnable parameters. 

\textbf{Action Spaces for Robot Learning.} Robot learning research has a rich history in combining well-established control methods -- often in the form of an action space -- with machine learning~\cite{kober2013reinforcement}. This hybrid approach can result in an improvement in sample efficiency, robustness, and quality of the learned policies~\cite{peng2017learning, varin2019comparison}. In \cite{schaal1994robot}, a dual-arm system learns to juggle by learning a world model and an optimal control policy. \cite{peters2007reinforcement} use an Operational Space based action space to learn an RL policy that follows a reference acceleration. Another popular action representation is dynamical movement primitives (DMPs)~\cite{schaal2006dynamic} which utilize a low-dimensional parameterization of a dynamical system to generate smooth trajectories. DMPs are often used with policy search methods~\cite{kober2014policy} and HRL~\cite{daniel2012hierarchical, end2017layered}. ~\cite{buchli2011learning} propose to learn a variable impedance controller in joint space and DMPs to optimize energy consumption while refining an initial trajectory. Similar to our approach, ~\cite{martin2019variable} uses a hierarchical structure to combine a model-free high-level policy with an impedance controller. In contrast, we use a bio-inspired adaptive low-level policy which independently optimizes trajectory deviation, stability, and energy consumption. For manipulation, the exerted force of the robot to the environment is essential and some action spaces include explicit force control~\cite{kalakrishnan2011learning, beltran2020learning} or implicit modulation based on impedance control~\cite{martin2019variable,9146673}. In this work, the agent is equipped with direct means to apply a wrench to the environment. 

\section{Preliminaries}
\label{sec:background}
\textbf{Compliant Robot Control in RL Action Spaces.} Compliant robot behavior is a classical problem in robotics~\cite{hogan1984impedance, khatib1987unified}. It is most relevant when the goal of the robot is to manipulate its environment or the environment is only partially known~\cite{ott2008cartesian}. In cartesian impedance control, the inputs to the controller are typically the desired end-effector position $\bm{x}_d$ and velocity $\bm{\dot{x}}_d$, from which the controller computes joint torques $\bm{\tau}_u$. The torques are computed to realize a desired, compliant dynamic relationship between the robot motion and external forces~\cite{ott2008cartesian}.

Consider the dynamical model of the robot
\begin{equation}
    \bm{M}(\bm{q}) \bm{\ddot{q}} + \bm{C}(\bm{q}, \bm{\dot{q}}) \bm{\dot{q}} + \bm{g}(\bm{q}) = \bm{\tau}_u + \bm{\tau}_{\text{ext}},
\end{equation}
where $\bm{M}(\bm{q})$ is the symmetric, positive definite inertia matrix, $\bm{C}(\bm{q}, \bm{\dot{q}})$ is the Coriolis/centrifugal matrix, $\bm{g}(\bm{q})$ is the vector of gravity torques, $\bm{\tau}_u$ is the vector of joint torques and $\bm{\tau}_{\text{ext}}$ is the vector of external torques. Following the operational space formulation~\cite{khatib1987unified}, the control law of the impedance controller in task space coordinates is
\begin{align}
    \bm{\tau}_u(t) &= \bm{J}(\bm{q})^\mathsf{T} \bm{F}_u(t) , \\
         &= \bm{J}(\bm{q})^\mathsf{T} \big(- \bm{K} \bm{e}(t) - \bm{D} \bm{\dot{e}}(t)\big) ,
    \label{eq:impedance_classical}
\end{align}
with input wrench $\bm{F}_u(t)$, Jacobian matrix $\bm{J}(\bm{q})$, pose error $\bm{e}(t) = \bm{x}(t) - \bm{x}_d(t)$ and velocity error $\bm{\dot{e}}(t) = \bm{\dot{x}}(t) - \bm{\dot{x}}_d(t)$. The stiffness $\bm{K}$ and damping $\bm{D}$ matrices are both positive definite, symmetric matrices. For clarity, we drop the vector of gravity torques $\bm{g}(\bm{q})$. A full derivation can be found in~\cite{ott2008cartesian}. 

This controller can be used as an action space by tracking a reference signal generated by an RL agent. The action space $\mathcal{A}$ consists of absolute or relative values $\bm{x}_d(t) \in SE(3)$ and needs a fixed mechanical impedance $\bm{K}, \bm{D}$ to be implemented. By abuse of notation, we write $u(t)=f(a=\bm{x}_d)$ to indicate that the action generated by the agent is a time depended pose $\bm{x}_d(t)$. 

A reasonable extension to the fixed impedance action space is to equip the agent with the means to vary the impedance by extending the action space~\cite{martin2019variable}. This variable impedance action space consists of tuples $u(t) = f(a = (\bm{x}_d, \bm{K}_d, \bm{D}_d))$ and is implemented in a similar fashion. 

\textbf{Reinforcement Learning.} In this work, we construct the high-level policy by maximization of future task reward. The goal is to learn a policy to perform tasks in continuous action and state spaces. Therefore, the tasks are modeled as finite-horizon, discounted Markov Decision Processes (MDP) defined by the tuple $(\mathcal{S}, \mathcal{A}, \mathcal{P}, r, \gamma, \rho, T)$. Here, $\mathcal{S}$ is a continuous state space, $\mathcal{A}$ is a continuous action space, $\mathcal{P}: \mathcal{S} \times \mathcal{A} \rightarrow \mathcal{S}$ are the transition dynamics of transitioning between states given an action, $r : \mathcal{S}\times\mathcal{A}\rightarrow \mathbb{R}$ is a scalar field quantifying the reward, $\gamma \in [0,1)$ a discount factor, $\rho$ an initial state distribution and horizon $T$. The goal of the learner is to find a policy $\pi: \mathcal{S} \rightarrow \mathcal{P}(\mathcal{A)}$ which selects actions to maximize the expected future reward~\cite{sutton2018reinforcement}. Assuming, that the policy is a function parameterized by $\theta$, the policy is optimized such that
\begin{equation}
\theta^* = \mathop{\mathrm{argmax}}_\theta J(\theta) = \mathop{\mathrm{argmax}}_\theta E_{\pi} \left[ \sum_{t=0}^{T-1} \gamma^t r(s_t, a_t) \right].
\end{equation}

\section{Adaptive Force-Impedance Control Action Space}
\label{sec:method}
Consider the problem of learning a manipulation task with a robotic arm from observations and measurements. To succeed, the control policy needs to generate the correct high-level actions, which are relevant sub-goals required to complete the task. Additionally, it needs to translate these high-level actions to the actuation of the manipulator such that the sub-goals are achieved. To implement this behavior, we design an action space and policy according to the following principles:

\noindent \textbf{(P1)} \textit{Multimodal Observations} are at the core of most manipulation tasks. Visual, tactile, and proprioceptive sensing vary drastically in nature and temporal context. A control policy must process different modalities at different frequencies to make the most out of available data without unnecessarily increasing the complexity of decision-making and actuation.

\noindent \textbf{(P2)} \textit{System Stability} is required at all times while performing manipulation. Studies have shown that slow feedback delays can lead to undesirable dynamic behavior and in the worst case to unstable dynamics~\cite{hogan1984impedance, burdet2001central,  yang2011human}. Thus, we require the policy to be able to continuously adapt the interaction to stabilize the system dynamics.

\noindent \textbf{(P3)} \textit{Physical Efficiency} is usually not an explicit goal of reinforcement learning agents. One could add an energy-specific term to the reward function of the agent, however, increasing the number of factors in the signal might require careful tuning of the coefficients and can in some cases deteriorate the learning performance~\cite{faust2019evolving}. Hence, the policy should optimize its energy consumption independent of the reward signal.

\noindent \textbf{(P4)} \textit{Safety} is one of the most critical attributes of any intelligent agent. While negative reward is important to learn safe interaction, we want the robot to be as compliant as possible unless the task requires otherwise. This can prevent harming the system if unexpected collisions occur.

\noindent \textbf{(P5)} \textit{Learning Efficiency} is a central struggle in RL. Especially in robotics, the state and action spaces are continuous and can have large dimensions. The complexity increases exponentially with each extra dimension according to the curse of dimensionality~\cite{kober2013reinforcement}. To reduce the complexity of the learning problem, it is important to keep the dimensionality of the action space as low as possible.

\subsection{Simultaneous Learning and Adaption}
Based on these principles and inspired by studies in neuroscience~\cite{botvinick2009hierarchically, botvinick2012hierarchical, ikegami2021hierarchical}, we model the solution of a general manipulation task as a hierarchical control system that is capable of \textit{thinking fast and slow}. A similar concept has been explored in previous work~\cite{martin2019variable, allshire2021laser, botvinick2019reinforcement}. 

We factor the policy into two components. A \textit{slow} outer loop is modeled as a policy $\pi(o): \mathcal{O} \rightarrow \mathcal{A}$ which maps environment observations to actions. Its goal is to plan the correct actions to solve a given task and it is parameterized by $\theta$. We consider this loop to be analogous to slow neural feedback loops found in human and animal behavior used for deliberate decision-making~\cite{kawato1999internal, botvinick2012hierarchical}. 

In contrast to prior work~\cite{martin2019variable, allshire2021laser}, the \textit{fast} inner loop of the system is a second optimization process. In this work, we model it as a second policy that maps actions and plant measurements to robot control commands $f(a, y): \mathcal{A} \times \mathcal{Y} \rightarrow \mathcal{U}$.\footnote{In this work, we consider measurements to be real-time sensor data which is required for robot control and observations a compilation of this data and other exteroceptive sensor modalities, such as image information.} The inner policy continuously adapts the interaction and motion to minimize the trajectory error and consumed energy of the system. It is parameterized by its own set of parameters $\psi$ determining the physical behavior. The goal of the inner loop is to optimize the action commanded by the high-level policy. An illustration can be found in Figure~\ref{fig:graph_abstract}.

The central goal of this architecture is to incorporate fast and slow learning processes to reduce the computational complexity of the individual components. Intuitively, the high-level policy is not required to learn and adapt high-frequency physical properties of the interaction, such as contact stability or energy efficiency, consequently reducing the computational effort. 

\subsection{Cartesian Adaptive Force-Impedance Action Space} 
The goal of the low-level control policy is to compute the joint torques $\bm{\tau}_u$ to establish efficient, safe manipulation of the environment, such that the commanded actions of the high-level policy are successful. For the environments we are interested in, we require low target error, stable interaction during contact-rich actions, and explicit force control for environments where it is necessary.
\begin{figure*}[t!]    
    \centering
    \includegraphics[width=\textwidth]{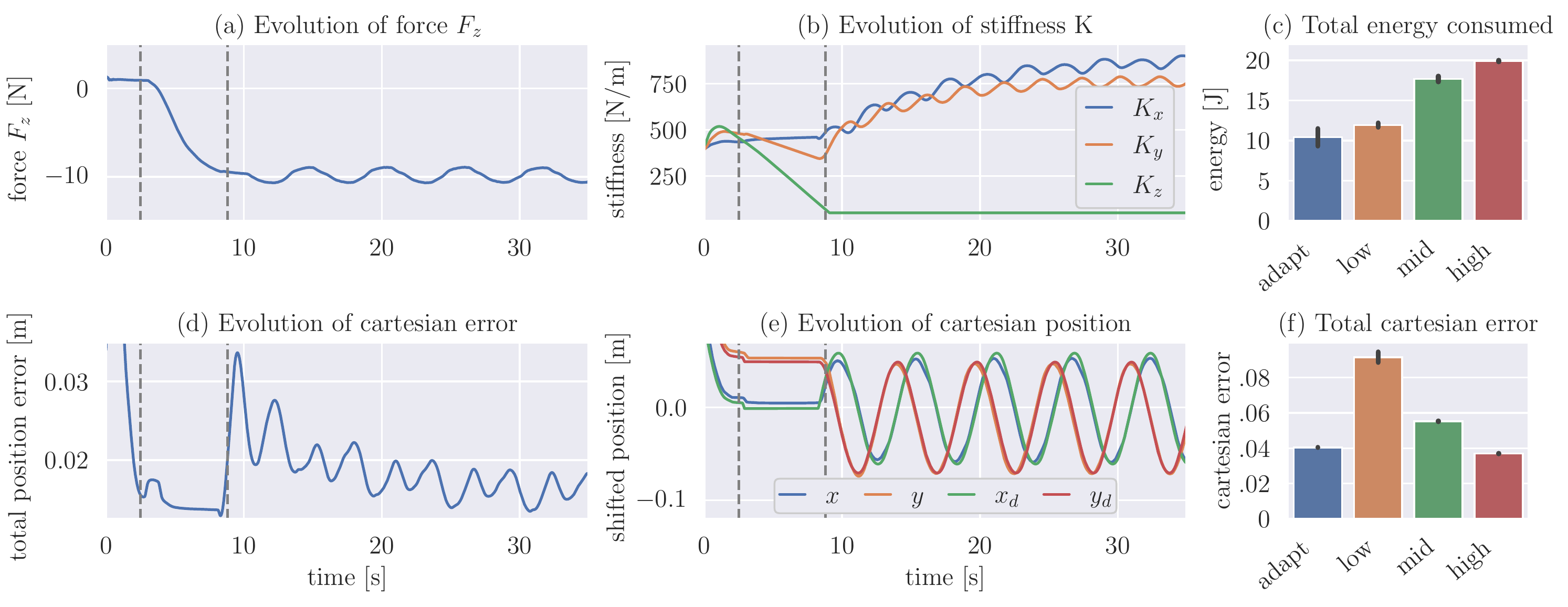}
    \caption{Evolution of the low-level control policy during a wiping task. (a) AFORCE allows precise application of force to the environment. (b) Once the wiping motion starts, the stiffness of the manipulator is not sufficient to keep a low tracking error, hence the policy increases it. (d) + (e) The evolution of the cartesian position and error clearly indicates the adaption to the new task. (c) + (f) In comparison to other compliant action spaces, AFORCE reduces the energy consumption while maintaining a competitive tracking error.}
    \label{fig:real_world}
\vspace{-4mm}
\end{figure*}

In this work, the low-level policy is a bio-inspired adaptive force-impedance controller~\cite{yang2011human}. Instead of varying the impedance through the neural feedback of the task planner, we adapt it through the minimization of instability, motion error, and effort~\cite{gomi1998task, franklin2008cns}. 

Following the formulation in~\cite{ganesh2010biomimetic, yang2011human}, we model the input wrench $\bm{F}_u(t)$ as a composition of a restoring feedback term $\bm{r}(t)$ (see (\ref{eq:impedance_classical})) and feedforward term $\bm{v}(t)=-\bm{F}_{ff}(t)-\bm{F}_d(t) $~\cite{johannsmeier2019framework}, such that
\begin{align}
\bm{\tau}_u(t) &= \bm{J}(\bm{q})^\mathsf{T} \big(\underbrace{- \bm{F}_{ff}(t) - \bm{F}_d(t)}_{\bm{v}(t)} \underbrace{-\bm{K}(t)\bm{e}(t) -\bm{D}(t)\bm{\dot{e}}(t)}_{\bm{r}(t)}\big).
\label{eq:aforce}
\end{align}
Here, $\bm{F}_{ff}(t)$ denotes an adaptive feedforward wrench that optimizes the motion of the manipulator~\cite{franklin2008cns}. An optional desired wrench profile $\bm{F}_d(t)$ can be added for environments that require specific force control, such as wiping or many other tool applications.

We adapt the impedance and feedforward wrench according to
\begin{align}
    \label{eq:stiffness_adaption}
    \bm{\dot{K}}(t) &=\bm{\beta} |\bm{\epsilon}(t)| - \bm{\gamma} \bm{\mathbb{1}} \\ 
    \label{eq:feedforward_adaption}
    \bm{\dot{F}}_{ff}(t) &= \bm{\alpha} \bm{\epsilon}(t) - \bm{\mu} \bm{F}_{ff}(t)
\end{align} 
with feedback error $\bm{\epsilon}(t) = \bm{e}(t) - \delta \bm{\dot{e}}(t)$ and $\delta > 0$. In addition, $\bm{\alpha}, \bm{\beta}, \bm{\gamma}, \bm{\mu}$ are constant symmetric positive-definite matrices that determine the adaption behavior of the control policy. $\bm{\alpha}$ and $\bm{\beta}$ are considered to be stiffness and feedforward adaption rates that determine the speed at which the controller can adapt to errors. In contrast, $\bm{\gamma}$ and $\bm{\mu}$ are relaxation factors that decrease the exerted energy. 

Intuitively, the first term of equation (\ref{eq:stiffness_adaption}) adapts the mechanical stiffness proportionally to the absolute error, reducing the tracking deviation. Simultaneously, the second term $- \bm{\gamma} \bm{\mathbb{1}}$ constantly decreases the stiffness, thus relaxing the arm and reducing the energy consumption along dimensions where the stiffness is currently not required. Hence, the impedance adaption law allows concurrent minimization of the energy and tracking error. A similar intuition can be employed for (\ref{eq:feedforward_adaption}).

The resulting impedance and feedforward wrench at time $t$ is thus $\bm{K}(t) = \int_0^t \bm{\dot{K}}(t) \diff t + \bm{K}(0)$, $\bm{D}(t)=2\sqrt{\bm{K}(t)}$ and $\bm{F}_{ff}(t) = \int_0^t \bm{\dot{F}}_{ff}(t) \diff t$, respectively. 

For clarity, we compile all adaptive parameters $\psi = (\bm{\alpha}, \bm{\beta}, \bm{\gamma}, \bm{\mu})$ and define the Cartesian \underline{A}daptive \underline{For}ce-Impedan\underline{ce} (AFORCE) action space as
\begin{equation}
    u(t) = f_\psi\big(a = (\bm{x}_d, \bm{F}_d)\big),
\end{equation}
with desired pose $\bm{x}_d(t)$ and optional wrench $\bm{F}_d(t)$. The values of $\psi$ can be chosen by a designer or learned. In this work, we choose them experimentally and leave learning them in a complete RL framework for future work.

AFORCE is agnostic to the choice of a task planner. A suitable policy can either be directly programmed through expert knowledge, learned from demonstrations or through reinforcement.

\section{Experiments}
\label{sec:experiments}
We evaluate the performance of AFORCE in both simulation and on real hardware. In simulation, we assess the learning performance on three challenging tasks using the \texttt{robosuite} simulator~\cite{zhu2020robosuite}. On real hardware, we demonstrate the efficacy of AFORCE on a contact-rich manipulation task. In the experiments, we aim to answer the following questions: (i) Is the action space suitable for learning robotic manipulation tasks on real hardware? (ii) How does the action space perform compared to baselines for contact-rich tasks? (iii) Is AFORCE suitable for model-free reinforcement learning? (iv) Does the action space enable safe and energy-efficient learning?
\begin{figure*}[!t]
    \centering
    \includegraphics[width=\textwidth]{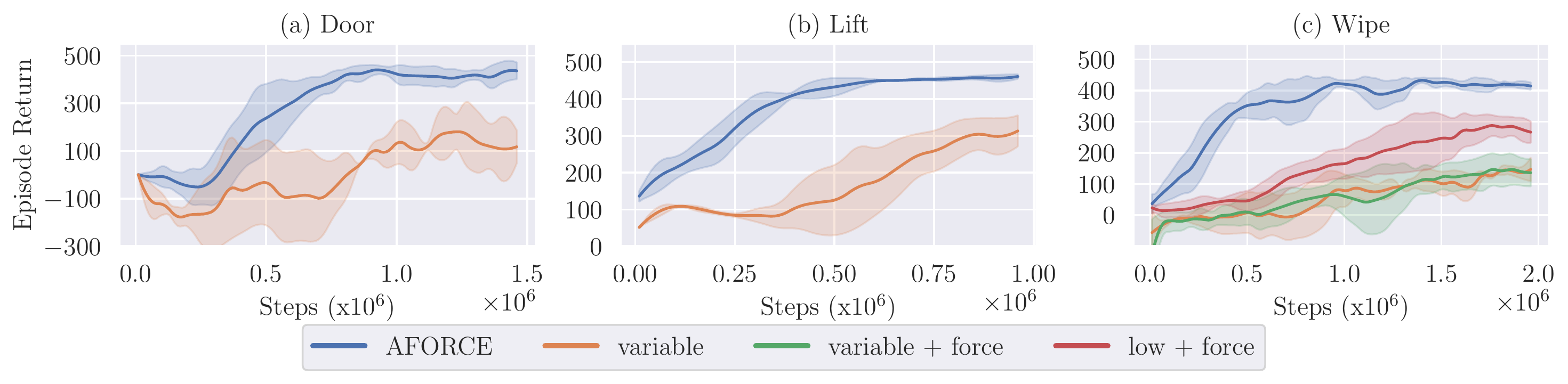}
    \caption{Training curves for the three tasks: \texttt{Door}, \texttt{Lift} and \texttt{Wipe}. The graphs depict mean and standard deviation of multiple different random seeds. In all tasks, AFORCE outperforms all alternatives and leads to more sample efficient and stable learning.}
    \label{fig:robosuite_results}
\vspace{-4mm}
\end{figure*}

\subsection{Real-World Experiments}
In all experiments, the high-level control policy outputs actions at $\SI{20}{\hertz}$, and the low-level policy runs at $\SI{1}{\kilo \hertz}$. We use a jerk-limited online trajectory generator to generate smooth trajectories from the actions~\cite{berscheid2021jerk}. We also show qualitative results of the robot in action in the supplementary video. 

\textbf{Contact-Rich Manipulation.} In this experiment, we demonstrate that the hierarchy and action space are suitable to solve contact-rich manipulation tasks and investigate the adaptive properties of the policy. To this end, we use an expert high-level policy to solve a wiping task. To highlight the multimodality of the task, we briefly cover its measurement and observation space. The measurement space at $\SI{1}{\kilo \hertz}$ for the experiment consists of robot joint positions and velocities, end-effector pose and velocity, and an estimated external wrench, such that the agent can control the force. The observation space at $\SI{20}{\hertz}$ consists of the current pose, recent external wrench measurements, and the centroid of the wipe spot, that can be detected with a visual sensor.

The evolution of the experiment is shown in Figure~\ref{fig:real_world}. After detecting the wipe spot, the policy computes actions to approach the centroid and establishes contact (first dashed vertical line). Next, it commands to apply a force of $\SI{10}{\newton}$ to the surface. Once the desired force is applied (second dashed vertical line), the policy commands a circular motion perpendicular to the surface via the action space while trying to maintain the same amount of force. The adaption of the policy to optimize the motion can be seen in (b), (d), and (e). During the time in which the policy applies the specified force, the mechanical impedance of the system decreased due to a low cartesian error. However, once the circular motion commences, the current impedance is not sufficient to achieve a low tracking error. Thus, it increases the stiffness periodically to adapt to the new motion. The result of the adaption to decrease the total cartesian tracking error is shown in (d) and (e).

\textbf{Energy Efficiency and Tracking Error.} In addition, we investigate how AFORCE performs in contrast to other compliant action spaces in a wiping task. In particular, we compare AFORCE to three fixed impedance action spaces with a \textit{low}, \textit{mid}, and \textit{high} impedance configuration. All spaces use the same force controller. 

We employ the same task plan as in the previous experiment but extend the wiping phase to wipe a total of eight full rotations. To acquire representative results, we conduct the same experiment for multiple trials. The efficiency of the different action spaces is evaluated by comparing the total energy consumption of the manipulator and the total tracking error\footnote{sum of absolute position error and quaternion difference} along the episode (see Figure~\ref{fig:real_world}). 

All action spaces except \textit{low} managed to wipe off all the paint. We found that AFORCE only consumes about half the energy of \textit{high} while maintaining a competitive total error. In our experiments, \textit{low} did not manage to maintain full contact in its trials, thus leaving paint unwiped. 

\subsection{Simulated Experiments}
In simulation, we conduct experiments on three manipulation tasks using the \texttt{robosuite} simulator~\cite{zhu2020robosuite}: \texttt{Door}, \texttt{Lift} and \texttt{Wipe}. The chosen tasks aim to cover a wide range of common manipulation challenges, such as contact dynamics, and kinematic constraints. We use Soft Actor Critic (SAC)~\cite{haarnoja2018soft, pytorch_sac}, as our RL policy, which is a model-free, off-policy deep RL algorithm. The actor and critic both consist of two layers with $1024$ units each. To evaluate the performance of AFORCE, we train multiple different seeds for each task and action space. At the beginning of each seed, we pretrain the models with $5$ thousand samples collected by rolling out random actions. Then, we evaluate the model every $10$ thousand steps for $10$ episodes and report the average reward.
\begin{table}[htb!]
\caption{Comparison of energy consumption and task performance of the simulated tasks.}
\centering
\begin{tabular}[t]{lcccccccc}
\toprule[1pt]
& \multicolumn{3}{c}{Joule / Action}  &\multicolumn{3}{c}{Reward} \\ 
\cmidrule(lr){2-4} \cmidrule(lr){5-7}
 &Door &Lift &Wipe  &Door &Lift &Wipe \\
\cmidrule(lr){1-7}
AFORCE              &\textbf{7.0}    &\textbf{1.5}    &1.8  &\textbf{456$\pm$20}  &\textbf{480$\pm$1} &\textbf{437$\pm$24}\\
\textit{var}            &131.0  &24.5   &27.0  &242$\pm$227 &363$\pm$66 &178$\pm$38 \\
\textit{var force}   &--     &--     &19.5  &--     &-- &178$\pm$48 \\
\textit{low force}         &--     &--     &\textbf{1.0}  &--     &-- &264$\pm$73 \\
\bottomrule[1pt]
\end{tabular}
\label{tab:energy_performance} 
\end{table}

\textbf{Tasks.} In \texttt{Door}, the agent has to learn to open a door by pushing down a spring-loaded handle and pulling the door open. We penalize excessive contact forces over $\SI{60}{\newton}$. In \texttt{Lift}, the goal of the agent is to grasp an object and lift it off the table. Lastly, in \texttt{Wipe}, the agent has to learn to wipe markers off a table using a wiping tool and applying a force above $\SI{15}{\newton}$. We add the same penalty as in \texttt{Door} and only remove markers off the table if the force is below the force penalty threshold. Additionally, we vary the task and robot initialization on each task reset. The joints of the robot are initialized with Gaussian noise, in \texttt{Lift} the object position is varied, in \texttt{Door} the position and orientation of the door is varied and in \texttt{Wipe} the wipe spots are randomized.

\textbf{Low-level Control Policies.} In all simulated experiments, the high-level control policy outputs actions at $\SI{20}{\hertz}$. The low-level control policies run at $\SI{500}{\hertz}$, using a linear interpolator to ensure smooth trajectories generated from the high-level actions. For the comparison, we are using the \texttt{variable\_kp} implementation of \texttt{robosuite} with $u(t)=f^{\text{vkp}}(a=(\bm{x}_d,\bm{K}_d))$ which is a \textit{variable} impedance action space. The environments \texttt{Door} and \texttt{Lift} do not require the explicit application of force, thus the AFORCE action space becomes $u(t)=f_\psi (a=\bm{x}_d)$. 

For the \texttt{Wipe} environment, we extend AFORCE by a desired wrench that is applied to the environment if contact is established. The policy is $u(t)=f_\psi(a=(\bm{x}_d,\bm{F}_d))$ and we implement the desired wrench using a PID controller~\cite{schindlbeck2015unified}. The controller can apply the desired wrench up to $\SI{50}{\newton}$ but not enough to cause a penalty. To gain insights on the importance of a force controller, we also evaluate the performance of \textit{variable + force} $u(t)=f^{\text{vkp}}(a = (\bm{x}_d, \bm{K}_d, \bm{F}_d))$. Similarly, we evaluate \textit{low + force}, a low impedance action space with force control $u(t)=f^{\text{low}}(a = (\bm{x}_d, \bm{F}_d))$. 

\textbf{Results: Learning Performance.} The results of the simulated experiments are shown in Figure~\ref{fig:robosuite_results}. In our comparison against baselines, AFORCE outperforms alternatives on all tasks. For all three tasks, it converges faster and reaches a higher maximum reward than the other action spaces.

In \texttt{Door}, AFORCE achieves a higher maximum reward and lower standard deviation than \textit{variable}. We attribute that to the lower average stiffness of AFORCE. Hence, the collision penalties are smaller than of an agent which explores its stiffness. The percent of episodes with a penalty is twice as high for \textit{variable} than for AFORCE and the average penalty is smaller. However, these values depend drastically on the success of the learning process, thus they should be merely seen as an indicator.

In \texttt{Lift}, we observe a similar result. \textit{variable} requires around $300$ thousand steps before it starts to converge. Since there are no negative rewards in this environment, a contributing factor may be the increased size of the action space which needs to be explored. Furthermore, AFORCE achieves a very stable reward during evaluation across multiple seeds.  Once the task is learned, the reward variance of AFORCE in \textit{Lift} becomes small and all policies succeed in most evaluation episodes.

In \texttt{Wipe}, we compare AFORCE to three action spaces, with and without force control. The task demands precise force application to wipe the dirt spots. We observe that the performance of \textit{variable} is almost identical with explicit force control. During the training period, the stiffness of \textit{variable} is higher than that of the other action spaces. The high stiffness resulted in high external forces when the manipulator made contact. This made it impossible for the force controller to regulate the desired force. Even when the \textit{variable + force} agent chose the right desired force, the controller was unable to stabilize the interaction. The \textit{low + force} space performs better than the \textit{variable} spaces. It wipes on average more markers but cannot consistently remove all. After inspecting training videos, it is often seen sliding on the table without making full contact. We attribute this to the fact that the force controller is engaged when the tool makes contact, but the agent is unable to align the tool with the table and stabilize the interaction. At last, AFORCE reaches its maximum reward consistently after $500$ thousand steps. We observe that AFORCE manages to wipe off all markers after only a few training episodes. Our action space can align the tool with the surface to use the force controller and simultaneously stabilize the interaction at the contact points.

\textbf{Results: Energy Efficiency and Safety.} First, we compare the average energy one action consumed during training (see Table~\ref{tab:energy_performance}). We found that \textit{variable} spaces consumed more energy in all three environments. In the \texttt{Wipe} environment, the \textit{low + force} action space needed the least amount. We observe that \textit{variable + force} used less energy on average than \textit{variable}. This could mean that the force controller attempts to regulate the measured external forces to the desired, safe value but fails. However, to definitively answer this question, more experiments are required.

Finally, we investigate how safe the learning of the different policies is. Hence, we evaluate two metrics: (a) the percentage of training episodes with a penalty, and (b) the average penalty per episode. We observe that no action space was completely safe. AFORCE performs best in terms of metric (a), yet received a penalty in $12\%$ of episodes in \texttt{Wipe}. In our experiments, the \textit{variable} spaces did not converge to a physically feasible solution. They are penalized in more than $75\%$ of the episodes and collect a significant amount of negative reward which is scaled by the amount of force above the penalty threshold. For (b), in \texttt{Wipe} the force penalty of the \textit{low + force} was the lowest, with $17.16$ negative reward per episode, closely followed by AFORCE with $17.76$. In \texttt{Wipe}, both \textit{variable} action spaces collect on average above $400$ penalty per episode. In \texttt{Door}, AFORCE collects on average $25$ penalty and \textit{variable} $149$.

\section{Conclusion and Future Work}
We presented AFORCE, an action space for learning contact-rich manipulation tasks. It draws inspiration from neuroscience and biomechanics to allow concurrent optimization of the task strategy, physical interaction, and motion. We demonstrated our approach on real hardware, showed its ability to solve a complex, dynamic manipulation task, and compared its energy efficiency to other compliant action spaces. We also showed that our method outperforms action space alternatives in simulation in terms of sample efficiency, learning stability, energy consumption, and safety. The main limitation of this work is that the adaptive parameters have to be tuned manually. While there is a large set of valid parameters, an invalid set can have a drastic impact on the learning performance. Additionally, the adaption of the interaction is reactive and the agent has no means to adapt it proactively. For future work, we plan to address these problems by learning the adaptive parameters and high-level policy concurrently and investigate methods that allow the agent to interfere with the adaption behavior.

\section*{Acknowledgement}
We gratefully acknowledge the general support by Microsoft Germany (Microsoft Deutschland GmbH), the funding of this work by the Deutsche Forschungsgemeinschaft through the Gottfried Wilhelm Leibniz Programme (award to Sami Haddadin; grant no. HA7372/3-1), and the funding of the Lighthouse Initiative Geriatronics by StMWi Bayern (Project X, grant no. 5140951) and LongLeif GaPa gGmbH (Project Y, grant no. 5140953). Please note that S. Haddadin has a potential conflict of interest as shareholder of Franka Emika GmbH.

\bibliographystyle{plain}
\bibliography{bibliography}

\addtolength{\textheight}{-12cm}   

\end{document}